\documentclass{article}

\usepackage{PRIMEarxiv}
\usepackage[dvipsnames]{xcolor}
\usepackage[utf8]{inputenc} 
\usepackage[T1]{fontenc}    
\usepackage{hyperref}       
\usepackage{url}            
\usepackage{booktabs}       
\usepackage{amsfonts}       
\usepackage{nicefrac}       
\usepackage{microtype}      
\usepackage{lipsum}
\usepackage{fancyhdr}       
\usepackage{graphicx}  
\usepackage{amsmath}      
\usepackage{algorithm}
\usepackage{algorithmic}
\graphicspath{{media/}}
\usepackage{multirow}
\usepackage{float}
\usepackage{color, colortbl}
\usepackage{authblk,amssymb}
\definecolor{Gray}{gray}{0.9}
\makeatletter
\renewcommand\AB@affilsepx{, \protect\Affilfont}
\makeatother
  
\title{ESPERANTO: Evaluating Synthesized Phrases to Enhance Robustness in AI Detection for Text Origination
}

\author[1, 4]{Navid Ayoobi}
\author[4]{Lily Knab}
\author[4]{Wen Cheng}
\author[2,4]{David Pantoja}
\author[3,4]{Hamidreza Alikhani}
\author[4]{Sylvain Flamant}
\author[4]{Jin Kim}
\author[1]{Arjun Mukherjee}

\affil[1]{University of Houston}
\affil[2]{University of California, Berkeley}
\affil[3]{University of California, Irvine}
\makeatletter
\renewcommand\AB@affilsepx{ \protect\Affilfont}
\makeatother
\affil[4]{Esperanto Technologies}
\affil[ ]{\texttt{nayoobi@cougarnet.uh.edu}}

\newcommand{\mycompany}{Esperanto AI}
\begin{document}
\maketitle
\vspace{-35pt}
\begin{center}
    \text{\mycompany}
\end{center}
\vspace{35pt}

\begin{abstract}
While large language models (LLMs) exhibit significant utility across various domains, they simultaneously are susceptible to exploitation for unethical purposes, including academic misconduct and dissemination of misinformation. Consequently, AI-generated text detection systems have emerged as a countermeasure.However, these detection mechanisms demonstrate vulnerability to evasion techniques and lack robustness against textual manipulations. This paper introduces back-translation as a novel technique for evading detection, underscoring the need to enhance the robustness of current detection systems.
The proposed method involves translating AI-generated text through multiple languages before back-translating to English. We present a model that combines these back-translated texts to produce a manipulated version of the original AI-generated text. Our findings demonstrate that the manipulated text retains the original semantics while significantly reducing the true positive rate (TPR) of existing detection methods.
We evaluate this technique on nine AI detectors, including six open-source and three proprietary systems, revealing their susceptibility to back-translation manipulation.
In response to the identified shortcomings of existing AI text detectors, we present a countermeasure to improve the robustness against this form of manipulation.
Our results indicate that the TPR of the proposed method declines by only $1.85\%$ after back-translation manipulation.
Furthermore, we build a large dataset of 720k texts using eight different LLMs.
Our dataset contains both human-authored and LLM-generated texts in various domains and writing styles to assess the performance of our method and existing detectors. This dataset is publicly shared for the benefit of the research community.

\end{abstract}


\section{Introduction}

Through the training on a substantial volume of textual data, large language models (LLMs) encapsulate knowledge across various fields, incorporate a range of writing styles, and maintain contextual comprehension within their parameters.
This has rendered them ubiquitous and state-of-the-art in numerous applications like translation \cite{feng2024improving}, summarization \cite{zhang2024systematic}, text classification \cite{milios2023context}, chat bots and virtual assistants \cite{kim2023chatgpt}.
With their ability to be prompted effortlessly at minimal cost and their proficiency in generating high-quality and human-like text, LLMs are becoming an attractive tool for malicious users.
Malicious activities include, but are not limited to, academic dishonesty \cite{grande2024student}, the production of fake news \cite{su2023fake}, scam messages \cite{ayoobi2023looming}, fraudulent reviews \cite{salminen2022creating}, and automated cyberbullying \cite{verma2024beyond}.
Beyond deliberate misuse, there are instances where LLMs inadvertently generate outdated information \cite{chen2023combating}, such as within a question-answering framework, due to training on obsolete data.
LLMs are also prone to producing hallucinations \cite{ji2023towards,yao2023llm}, which are convincingly realistic yet factually incorrect or nonsensical information.
In addition, given the pervasive presence of AI-generated content, it is occasionally essential to distinguish and filter out human-generated data from contaminated training sets for effective machine learning model training.
Hence, it is crucial to establish a clear distinction between human-written and AI-generated contents to prevent potential misinformation issues.

Currently, humans exhibit a moderate ability to distinguish AI-generated content, which is nearly equivalent to a random classifier \cite{gehrmann2019gltr,clark2021all,perkins2023game,uchendu2021turingbench,dou2021gpt,soni2023comparing}. 
This can be attributed to the fact that humans are unable to detect recurring patterns and universal traits among all AI-generated texts.
As a result, the close resemblance between AI-generated text and human-authored text poses serious challenges for identification by humans \cite{ayoobi2024seeing}.
In order to mitigate these risks, several studies offer cues to enhance people's ability to distinguish AI-generated text \cite{ayoobi2024seeing,munoz2024contrasting,georgiou2024differentiating}. 
However, these efforts have limited practical effectiveness since educating all individuals is a nearly impossible task.
Furthermore, with the rapid evolution of LLMs, the cues may not remain effective in all scenarios.
Therefore, a more viable alternative is to develop and implement automatic detectors for AI-generated text.
These detectors generally treat AI content detection as a binary classification task, and classify a piece of text as either generated by an LLM or composed by a human \cite{zhu2023beat}.
In this context, a trade-off exists between the false positive rate (incorrectly labeling human-written content as AI-generated) and the false negative rate (misclassifying AI-generated content as human-written).
The former can be exploited by adversaries through spoofing attacks, damaging the reputation of LLM developers \cite{sadasivan2023can}, while the latter, if high, renders an AI detector ineffective.
In this paper, our main focus is on enhancing robustness to counter evasion from detection.

Scholars have pointed out that the robustness of current AI detection methods is uncertain as they struggle with several issues including domain-specific  dependencies, dependence on generator models \cite{bethany2024deciphering}, out-of-distribution scenarios \cite{wu2023survey}, and bias against non-native English speakers \cite{hu2023radar,emi2024technicalreportpangramaigenerated}.
Inspired by digital watermarking techniques used in multimedia, numerous studies have designed AI-text detectors that utilize the insertion of hidden patterns within AI-generated text to assist in identifying and confirming the origin of the content \cite{kirchenbauer2023watermark,kirchenbauer2023reliability,hou2023semstamp,liu2023private}.
However, text manipulations like paraphrasing can compromise the robustness of watermarking methods \cite{hu2023radar,sadasivan2023can,krishna2024paraphrasing}.
The reality is that introducing additional manipulation techniques to evade detection, and subsequently proposing potential solutions, aids in paving the way towards more robust approaches. 
In this paper, we explore the impact of back-translation for the first time as a text manipulation technique in circumventing current AI-text detection methods, and then presents a method to counteract its negative consequences.

Back-translation has been extensively utilized for data augmentation, especially for low-resource languages in neural machine translation (NMT) systems \cite{fadaee2017data,hemmatizadeh2023latent,gao2023data}.
This technique plays a pivotal role in enabling  multilingual NMT models to improve the translation of low-resource language pairs and enable zero-shot translation automatically and without additional data augmentation \cite{zhang2020improving,zhang2023improving}. 
In this paper, we introduce back-translation as a technique for manipulating AI-generated text to bypass detection.
The primary distinction between back-translation employed in NMT models and utilized in our work lies in the fact that in NMT, the source text is initially translated to an intermediate language and then the outcome is translated to the target language.
In contrast, our approach involves translating the AI-generated English text to an arbitrary language (other than English) and subsequently back-translating the result to English.
After generating equivalent back-translated texts from various intermediate languages, we combine them using our proposed method based on the word error rate (WER) metric to construct a manipulated version of the AI-generated text. 
We demonstrate that the combined text degrades the performance of existing AI-text detection methods.
For instance, the true positive rate (TPR) of RADAR \cite{hu2023radar} drops by $52\%$ on a question-answering dataset following the application of back-translation.
We ensure that the combined text preserves the same semantics as the AI-generated text by measuring the similarity between the combined and AI-generated texts using two different similarity measures \cite{wieting2021paraphrastic,cer2018universal}.
In addition, we created a large-scale dataset comprising human-authored and LLM-generated text samples in multiple writing styles, including journalistic, scientific, informative and everyday writing across three different proficiency levels. 
We release this dataset to the research community to aid in advancing further investigations in this subject.
Based on our experiments and results, this dataset already challenges existing detection methods, even prior to the application of our proposed back-translation technique.
Our results further indicate that our proposed method intensifies the challenges for existing detection systems, raising concerns about their robustness. 
We utilized eight LLMs to create our dataset: Mistal-7b \cite{jiang2023mistral}, Llama3-8b \cite{llama3}, Llama3-70b \cite{llama3}, GPT 3.5 Turbo \cite{gpt}, Phi3-Medium \cite{abdin2024phi}, Yi-34b \cite{young2024yi}, Llama3.1-8b \cite{dubey2024llama3herdmodels}, and GPT4o-mini\cite{gpt}. 
By manipulating solely the outputs of these LLMs, we illustrate that our proposed back-translation manipulation can evade detection without requiring white-box access to the architecture of LLMs or detection models.
We evaluate the effectiveness of our method in evading detection on nine different open-source and commercial AI-text detectors. 
The open-source methods analyzed in this study include RADAR \cite{hu2023radar}, LLMDet\cite{wu2023llmdet}, Likelihood, Rank, Log-Rank, and ESAS\cite{ayoobi2024seeing}, while the proprietary models tested are Pangram\cite{emi2024technicalreportpangramaigenerated}, GPTZero \cite{gptzero}, and ZeroGPT \cite{zerogpt}.

Additionally, to counter the effects of back-translation manipulation, we present a detection technique specifically designed to withstand this form of manipulation.
We demonstrate that the proposed method experiences a mere $1.85\%$ reduction in TPR when AI-generated text undergoes back-translation manipulation.

Our primary contributions can be outlined as follows:
\begin{itemize}
    \item We design and build a large dataset comprising 720k human-authored and LLM-generated texts in multiple writing styles, challenging the robustness of existing methods. We release this dataset to the research community.
    \item We propose a novel text manipulation technique based on back-translation, which can effectively evade current AI-text detectors. This finding raises concerns about the robustness of these detection methods.
    \item We devise a countermeasure to address the manipulated text produced by back-translation, thereby enhancing the robustness of AI-text detection systems against such breaches.
\end{itemize}

\section{Related work}
The identification of machine-generated text has been an active field of study preceding the unveiling of LLMs \cite{nichols2009machine,labbe2013duplicate}.
The emergence of LLMs has heightened the urgency and priority of devising effective techniques for the identification of synthetic content.
Broadly speaking, AI-text detection techniques can be classified into four categories: statistical \cite{gehrmann2019gltr,mitchell2023detectgpt,liang2024monitoring}, information retrieval \cite{krishna2024paraphrasing,li2024re,liao2023muser}, zero-shot \cite{zhu2023beat,bhattacharjee2024fighting}, and watermarking \cite{kirchenbauer2023watermark,kirchenbauer2023reliability,hou2023semstamp,liu2023private,jovanovic2024watermark} methods.
Statistical methods involve analyzing the distribution of linguistic patterns in a text to extract statistical features, which are subsequently used to determine whether the text is human-written or AI-generated. 
Building on the fact that most language models tend to sample from the head of the distribution, Gehrmann \textit{et al.} \cite{gehrmann2019gltr}. introduce a statistical approach that incorporates three tests: the probability of the word, the absolute rank of a word, and the entropy of the predicted distribution. 
These tests enable them to quantify the likelihood that a generated word is derived from the top of the distribution and to evaluate whether the previously generated context is recognized by the detection system.
The research conducted by Crothers \textit{et al. }\cite{crothers2022adversarial} demonstrates that despite the fact that neural network features outperform statistical features, the integration of statistical features can enhance the robustness against particular adversarial attacks.
By leveraging information retrieval principles, Krishna \textit{et al.} \cite{krishna2024paraphrasing} suggest a defense against paraphrase attacks through the retrieval of earlier-created AI-text. 
Their approach involves storing all LLM-generated texts in a database and then searching the entire database for a text that approximately matches the content of the input query.
However, retrieval-based detection methods require maintaining a substantial database of LLM-generated texts, and querying this database to find matches can be excessively time-consuming.

In an alternative approach to detecting AI-generated text, researchers have made attempts to utilize LLMs to compel them to identify the content that they have generated themselves in a zero-shot manner.
Based on the assumption that the ChatGPT \cite{gpt} model make fewer modifications to LLM-generated texts compared to human-written texts, Zhu \textit{et al.} \cite{zhu2023beat} develop a zero-shot and black-box detection method.
This approach generates revised versions of a text using ChatGPT and measures the similarity between the original text and its revised version. 
They use the criterion that a higher similarity score suggests a higher probability of the text being LLM-generated to assess whether a text is AI-generated.
In another research effort, Bhattacharjee and Liu \cite{bhattacharjee2024fighting} assess the zero-shot performance of ChatGPT by providing it with a simple prompt along with the text to be classified in the task of distinguishing between human-written and AI-generated text. 
 They test this approach on samples from 19 models, ranging from an 82M-parameter model to a 1.6B-parameter model, as found in the TuringBench dataset \cite{uchendu2021turingbench}. 
Their findings indicate that although ChatGPT has difficulty identifying AI-generated text, it performs effectively on human-written text.

To investigate the reliability of existing AI-text detectors, numerous studies have been dedicated to designing prompts that may allow LLMs to generate texts capable of evading detection. 
In one such work, Kumarage \textit{et al.} \cite{kumarage2023reliable} present a framework named ``EScaPe'', which directs pre-trained language models (PLMs) to circumvent AI-generated-text detectors using a universal evasive prompt.
The EScaPe framework involves initially crafting a specific evasive prompt for a particular PLM through prompt tuning and then capitalizes on the transferability of soft prompts to transfer the evasive prompt from one PLM to another.
In a related study, Lu \textit{et al.} \cite{lu2023large} propose ``SICO'', an in-context learning approach that iteratively replaces words and sentences within the in-context examples to assist LLMs in generating text that can evade detection. 
The substitution procedure is directed by a proxy detector.
The authors demonstrate that, in addition to reducing the effectiveness of existing AI text detectors, SICO decreases the likelihood of being recognized by humans.
Kirchenbauer \textit{et al.} \cite{kirchenbauer2023watermark} present a watermarking strategy designed to make synthetic text detectable even in short token spans. 
This method operates by generating a pseudo-random ``red'' list of tokens for each position in the sentence, where the ``red'' list generator is seeded with the prior token of that position only. 
A third party with access to the random number generator can recreate the red list for each token and count how often the red list rule is violated.
However, studies like \cite{hu2023radar,krishna2024paraphrasing} indicate that watermarking is vulnerable to text manipulations such as paraphrasing. 
As an example, Cai and Cui \cite{cai2023evade} reveal that a minor alteration, such as inserting a single space character before a random comma in AI-generated text, can deceive a detector.
To achieve robustness against paraphrasing, Hu et al. \cite{hu2023radar}, propose RADAR, which employs adversarial training to concurrently train a paraphraser and a detector in a two-player game scenario.
The role of paraphraser is to rephrase text from the training corpus in a way that diminishes the detector's likelihood of predicting it as AI-generated.
Conversely, the detector focuses on improving its detection capabilities by learning to compare human-written text with AI-generated text from both the training data and the paraphraser’s outputs.

Extending beyond previous studies, this paper introduces an innovative technique for manipulating AI-generated text that evades detection by existing detectors, including those designed to be robust against paraphrasing and other methods such as commercial ones.
We then present a method to counteract this manipulation to take a step forward in making AI-text detection more robust.

\section{Dataset}
For the purposes of this research, a large dataset was compiled, encompassing 72k instances of human-written texts and their corresponding AI-generated versions.
Additionally, a further 720k instances were generated from both human and AI-produced content via the technique of back translation, which will be detailed later in this section.
To ensure a diverse range of writing styles, our dataset includes four distinct text categories: news articles to represent journalistic style, paper abstracts to exemplify scientific style, Amazon product reviews to represent the informative review style, and responses to questions posted online to reflect the everyday writing style prevalent on the internet.

\textbf{News articles:} For the news articles, we utilized 3000 samples collected by Ayoobi et al. \cite{ayoobi2024seeing}.
We selected those generated articles produced by their \textit{``Summary Expanding''} strategy and the Mistral-7b model \cite{jiang2023mistral}.
These articles were originally sourced from reputable news agencies and subsequently converted into AI-generated counterparts following a process of summarization and expansion.
Detailed information about this pipeline is available in \cite{ayoobi2024seeing}.

\textbf{Paper abstracts:} A subset of 3000 scientific paper abstracts was sampled from the two million arXiv abstracts dataset introduced in \cite{clement2019use}. 
The Llama3-8b model \cite{llama3} was employed to identify the 10 most significant keywords and key phrases by providing the following prompt and a corresponding paper abstract: \textit{``You are a knowledgeable editor of a scientific journal trained to extract only 10 most important key words or phrases of a paper's abstract''}.
Additionally, we tasked the Llama3-8b model with summarizing each abstract into a single sentence by providing it with the prompt: \textit{``You are a knowledgeable editor of a scientific journal trained to summarize a paper's abstract in only one sentence with less than 30 words''}. 
After extracting the keywords and one-sentence summaries of the abstracts, we employed the Llama3-70b model \cite{llama3} to generate the AI counterpart for each abstract.
The model was guided by the prompt: \textit{``You are a knowledgeable scientific author trained to write a paper abstract containing [N] words given a list of key words and one-sentence summary of desired abstract''}.
we substituted \textit{[N]} with the original human-authored abstract's total word count to ensure length consistency.

\textbf{Reddit QA:} 
For the question and long-answer data, we collected questions from the \textit{``Explain Like I'm Five (ELI5)''} forum on Reddit, following a methodology similar to \cite{fan2019eli5}.
Initially, we filtered the questions to include only those with at least one answer exceeding 300 words.
Subsequently, 25k questions were randomly sampled from the filtered questions.
Five distinct LLMs, namely GPT 3.5 Turbo \cite{gpt}, Llama3-8b, Phi3-Medium \cite{abdin2024phi}, Mistral-7b, and Yi-34b \cite{young2024yi}, were employed to generate AI answers in three different proficiency levels: simplified, expert, and without any specific condition.
The respective prompts used for generating answers were: \textit{``Answer my question like I am five years old in about 300 words''}, \textit{``Answer my question like an expert in about 300 words''}, and \textit{``Answer my question in about 300 words''}.

\textbf{Product reviews:} 
To include shorter AI texts, we randomly selected 5000 product reviews, each containing 40 to 50 words, from five Amazon product categories as described in \cite{hou2024bridging}.
These categories included office products, fashion, pet supplies, health and personal care, and toys and games.
To generate AI counterparts, we first utilized the Llama3.1-8b model \cite{dubey2024llama3herdmodels} to extract three keywords from each review, prompted with: \textit{``You are an expert product reviewer trained to extract three most important keywords or phrases from a product review I give you''}.
Subsequently, from these 5000 reviews, we generated 2500 AI counterparts using the Llama3.1-8b model and another 2500 using the GPT4o-mini model.
The prompt used was: \textit{``You are an Amazon customer. You write a review about a product I give you in about [N] words. You must also use the keywords I give you in writing your review. The product is [PRODUCT] and the keywords are [KEYWORDS]''}.
In this prompt, \textit{[N]} was substituted with the total word count of the original human-authored review, \textit{[PRODUCT]} with the product title, and \textit{[KEYWORDS]} with the previously extracted keywords.

For both human-authored and AI-generated instances, we translated the text to an intermediate language and then back-translated it to English using Google Translate. We selected 10 different languages: Portuguese (PT), Spanish (ES), French (FR), Italian (IT), Chinese (ZH), Dutch (NL), Danish (DA), Japanese (JA), German (DE), and Korean (KO).
To maintain consistency, all texts (except reviews) were truncated to approximately 300 words. 
We observed that truncating mid-sentence decreases the detectability of AI-generated text.
Therefore, to maintain fairness in our dataset, we ensured that truncation occurred at the end of a complete sentence after the text's word count reached 300.
We refer to our dataset as \textbf{ESPERANTO}, which stands for \textbf{\underline{E}}valuating \textbf{\underline{S}}ynthesized \textbf{\underline{P}}hrases to \textbf{\underline{E}}nhance \textbf{\underline{R}}obustness in \textbf{\underline{A}}I detectio\textbf{\underline{N}} for \textbf{\underline{T}}ext \textbf{\underline{O}}rigination. This dataset is made publicly available for the research community\footnote{\url{https://github.com/navid-aub/Esperanto-Dataset}}

\section{Methodology}
In this section, we initially outline our proposed method of text manipulation through back-translation to evade detection.
We demonstrate that the manipulated text retains a high level of semantic similarity to the original AI-generated text using two distinct similarity metrics.
Furthermore, we detail a countermeasure to mitigate the impact of this manipulation on the robustness of a detection system.
\begin{figure*}
    \centering
    \includegraphics[width=\linewidth]{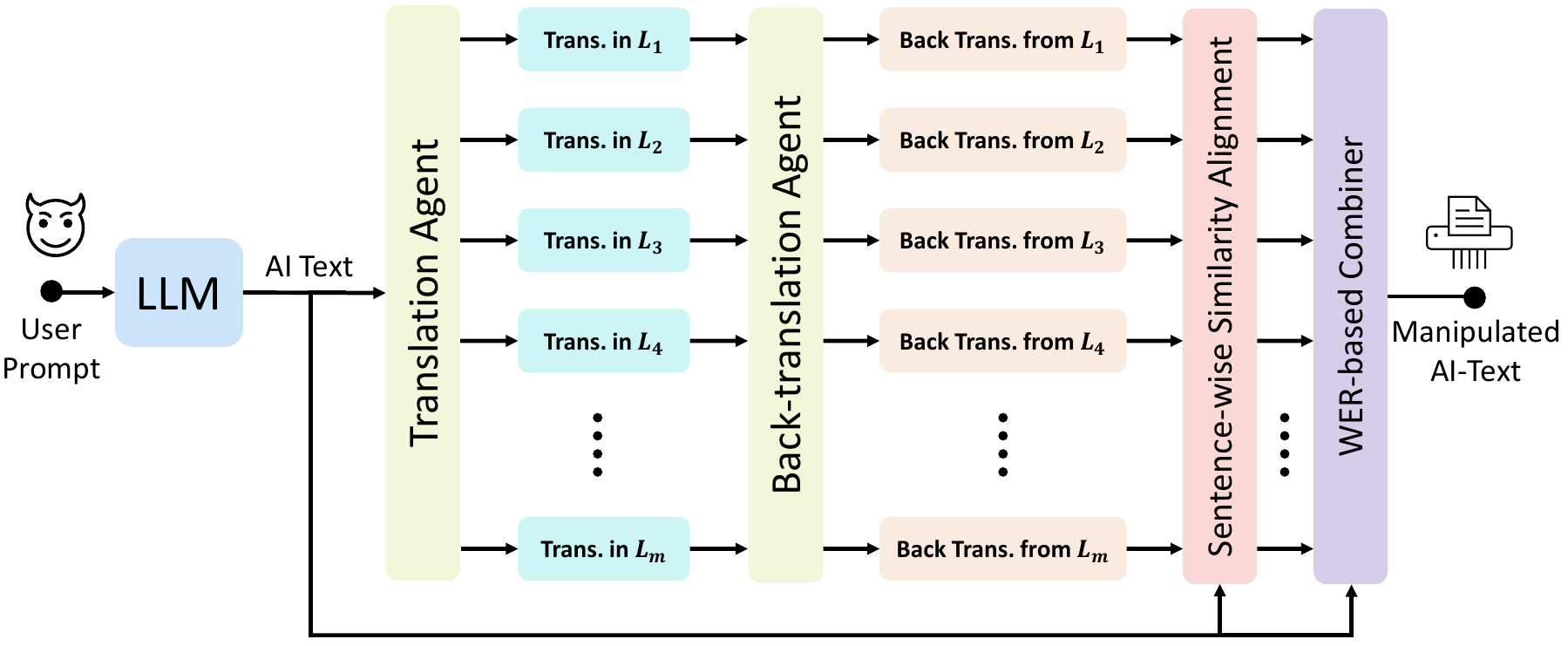}
    \caption{The overview of proposed method}
    \label{fig:overview}
\end{figure*}
\subsection{Manipulation of AI-generated text using back-translation}
Figure \ref{fig:overview} illustrates an overview of the proposed manipulation technique for evading detection. 
Initially, an LLM generates the desired content for a malicious user.
The AI-generated text $D^{Eng}_{0}$ is then translated into $m$ various languages (excluding English), indicated as $L_1, L_2, ...,$ and $L_m$, by a translation agent. 
Subsequently, a back-translation agent re-translates the text in language $L_j$ back into the original language (English), $D_{j}^{L_j}$.
We hypothesize that different languages may use synonyms or phrases that deviate from the original wording to maintain the same meaning.
Intermediate languages may also utilize distinct grammatical structures that leads to changes in sentence construction. 
In addition, the presence of idiomatic expressions or cultural references that lack direct translations may also necessitate the adoption of alternative phrasing when the text is back-translated.

To integrate back-translated texts, it is essential to identify semantically equivalent sentences in each back-translated text derived from different languages.
Initially, we tokenize the original text and back-translated text from language $L_j$ into individual sentences, $\{S^{Eng}_{0,1}, S^{Eng}_{0,2}, ..., S^{Eng}_{0,N_0} \}$ and $\{S^{L_j}_{j,1}, S^{L_j}_{j,2}, ..., S^{L_j}_{j,N_j} \}$, respectively. Where $N_0$ and $N_j$ indicate the total number of sentences in original AI text and the back-translated text from language $L_j$.
Then, for each sentence in the original text, we compute the similarity between that sentence and every sentence in a back-translated text, $\mathcal{\phi}$.
The sentence with the highest similarity score in $\mathcal{\phi}$ is subsequently designated as the corresponding sentence in the back-translated text.
This process is carried out in the ``sentence-wise similarity alignment'' block as illustrated in Figure \ref{fig:overview}.

To combine the selected sentences into a unified text, we employ the WER metric to compare the original sentence with the selected sentences from different back-translated texts, $\mathcal{\delta}$.
WER measures the discrepancy between two texts by calculating the number of substitutions, deletions, and insertions needed to transform the target text to match the reference text, normalized by the total word count of the reference text.
In this study, to increase the likelihood of evading detection, we select the sentence with the highest WER among the different back-translated texts in $\mathcal{\delta}$.
The concatenation of these sentences forms the manipulated text.
This procedure is performed by the ``WER-based combiner'' block depicted in Figure \ref{fig:overview}.
Algorithm \ref{algorithm} outlines the step-by-step procedure for combining back-translated texts based on the WER metric.

\begin{algorithm}[t]
    \caption{Combining back-translated texts based on word error rate}\label{algorithm}
    \begin{algorithmic}

    \STATE \textbf{Input:}
    \STATE An AI-generated text document $D^{Eng}_{0}$ 
    \STATE Sentence tokenized of $D^{Eng}_{0}$: $\{S^{Eng}_{0,1}, S^{Eng}_{0,2}, ..., S^{Eng}_{0,N_0} \}$
    \STATE A list of $m$ intermediate languages $L_j$'s
    
    \STATE \textbf{Preparation of back-translated documents:}
    \FOR {$j$ $=$ $1$ to $m$}
        \STATE $D_{temp}$ $\leftarrow$ Translate $D^{Eng}_{0}$ to language $L_j$
        \STATE $D_{j}^{L_j}$ $\leftarrow$ Translate $D^{temp}$ to English from language $L_j$
        \STATE Tokenize back-translated document $D^{j}_{L_j}$ into sentences: $\{S^{L_j}_{j,1}, S^{L_j}_{j,2}, ..., S^{L_j}_{j,N_j} \}$
    \ENDFOR
    
    \STATE \textbf{Combining back-translated documents:}
    \STATE Create an empty document $doc$  
    \FOR {$i$ $=$ $1$ to $N_0$}
        \STATE Create an empty WER array $\mathcal{\delta}$ with size $m$
        \FOR {$j$ $=$ $1$ to $m$}
            \STATE Create an empty similarity array $\mathcal{\phi}$ with size $N_j$
            \FOR {$k$ $=$ $1$ to $N_j$}
                \STATE $\phi$[$k$] $\leftarrow$ Compute the similarity between $S^{Eng}_{0,i}$ and $S^{L_j}_{j,k}$
            \ENDFOR
            \STATE $I_j$ $\leftarrow$ Select the index with maximum similarity in array $\mathcal{\phi}$
            \STATE $\mathcal{\delta}$[$j$] $\leftarrow$ Compute word error rate between $S^{Eng}_{0,i}$ and  $S^{L_j}_{j,I_j}$
        \ENDFOR
    \STATE $I_i$ $\leftarrow$ Select the index with maximum word error rate in array $\mathcal{\delta}$
    \STATE $doc$ $\leftarrow$ Concatenate $doc$ with $S^{L_{I_i}}_{I_i,I_j}$
    \ENDFOR

    \RETURN doc
    
    \end{algorithmic}
\end{algorithm}

\subsection{Evaluating similarity between AI-generated text, back-translated texts and combined text}
To confirm that the manipulated texts convey the same meaning as the original AI-generated texts, we employ two similarity measures, namely P-SP \cite{wieting2021paraphrastic}, and USEE \cite{cer2018universal}.
P-SP is a lightweight semantic similarity measure trained on over 25 million paraphrase pairs from the ParaNMT dataset \cite{wieting2017paranmt} using negative sampling.
It produces sentence embeddings by averaging the embeddings of sub-words within a sentence, as tokenized by SentencePiece \cite{kudo2018sentencepiece}. 
The similarity between two texts is reported by calculating the cosine similarity of their respective embeddings.
In the PAR3 dataset \cite{thai2022exploring}, human paraphrases yield an average P-SP score of $0.76$ \cite{krishna2024paraphrasing}.
In line with the methodology in \cite{krishna2024paraphrasing}, we regard semantics as approximately preserved if the P-SP score exceeds this average human paraphrase score. 

The Universal Sentence Encoder for English (USEE) is a deep averaging network-based sentence encoding model that leverages multitask learning to generate effective sentence representations.
Specifically, it calculates the mean of both word- and bi-gram-level embeddings, which are then passed through a feedforward deep neural network to produce sentence embeddings.
We adopted a similar procedure as described in \cite{krishna2024paraphrasing} to establish a threshold for semantic preservation between two texts using the USEE metric.
Accordingly, we calculated the USEE similarity between two translations derived from the same reference paragraph by two different human translators within the PAR3 dataset.
We considered the average USEE score of $0.69$ to be a critical threshold, above which semantic preservation was deemed to be approximately maintained.

We apply P-SP and USEE to compute the semantic similarity between the original AI-generated text and the back-translated texts derived from different languages.
We then use these metrics again to measure the semantic similarity between the original AI-generated texts and the final manipulated texts created by our proposed method.
The results of the former analysis are presented in Table \ref{tab:backtranslation_similarity}, while those of the latter are shown in Table \ref{tab:combined_similarity}.
The results consistently show that, across all intermediate languages and datasets, the metrics surpass their specified thresholds, approaching the maximal similarity value of 1, which confirms that the proposed method preserves the original semantics effectively after manipulation.

\renewcommand{\arraystretch}{1.1}
\begin{table*}[t]
\centering
\caption{The similarity between the original AI-generated texts and their back-translated versions for different languages}
\begin{tabular}{llccccccccc}
\hline
\rowcolor{Gray}\textbf{Language}& \textbf{Sim.} & \textbf{News} & \textbf{Abst.}& \textbf{ELI-G}& \textbf{ELI-L}& \textbf{ELI-M}& \textbf{ELI-P}& \textbf{ELI-Y}& \textbf{R-G}& \textbf{R-L} \\
\hline
\multirow{2}{*}{\textbf{Portuguese (PT)}} & P-SP &0.988&0.990&0.989&0.960&0.988&0.989&0.990 &0.964&0.964\\
                                          & USEE &0.987&0.984&0.987&0.959&0.987&0.988& 0.988&0.960&0.959\\     
\hline
\multirow{2}{*}{\textbf{Spanish (ES)}} & P-SP &0.988&0.988&0.988&0.961&0.988&0.989& 0.989&0.963&0.962\\
                                          & USEE &0.986&0.982&0.987&0.959&0.986&0.987&0.988&0.958&0.958 \\     
\hline
\multirow{2}{*}{\textbf{French (FR)}} & P-SP &0.985&0.987&0.986&0.963&0.986&0.986& 0.988&0.964&0.962\\
                                          & USEE &0.984&0.981&0.985&0.961&0.985&0.985& 0.987&0.960&0.959\\     
\hline
\multirow{2}{*}{\textbf{Italian (IT)}} & P-SP &0.988&0.989&0.988&0.963&0.989&0.989& 0.989&0.961&0.960\\
                                          & USEE &0.986&0.983&0.987&0.960&0.987&0.987& 0.988&0.956&0.955\\     
\hline
\multirow{2}{*}{\textbf{Chinese (ZH)}} & P-SP &0.970&0.972&0.976&0.938&0.974&0.977& 0.978&0.933&0.933\\
                                          & USEE &0.971&0.966&0.975&0.937&0.974&0.977& 0.978&0.933&0.932\\     
\hline
\multirow{2}{*}{\textbf{Dutch (NL)}} & P-SP &0.987&0.988&0.987&0.962&0.987&0.988& 0.989&0.965&0.963\\
                                          & USEE &0.986&0.981&0.986&0.961&0.986&0.987& 0.987&0.962&0.960\\     
\hline
\multirow{2}{*}{\textbf{Danish (DA)}} & P-SP &0.991&0.991&0.991&0.959&0.990&0.991& 0.992&0.972&0.973\\
                                          & USEE &0.989&0.987&0.990&0.958&0.988&0.990& 0.991&0.968&0.968\\     
\hline
\multirow{2}{*}{\textbf{Japanese (JA)}} & P-SP &0.967&0.972&0.969&0.932&0.971&0.973&0.973 &0.912&0.917\\
                                          & USEE &0.970&0.964&0.969&0.932&0.971&0.972& 0.973&0.920&0.920\\     
\hline
\multirow{2}{*}{\textbf{German (DE)}} & P-SP &0.983&0.984&0.983&0.959&0.982&0.984&0.985 &0.953&0.949\\
                                          & USEE &0.982&0.978&0.981&0.959&0.981&0.983&0.984 &0.951&0.947\\     
\hline
\multirow{2}{*}{\textbf{Korean (KO)}} & P-SP &0.973&0.973&0.974&0.935&0.973&0.975&0.976 &0.925&0.920\\
                                          & USEE &0.973&0.966&0.972&0.935&0.972&0.974& 0.975&0.926&0.920\\     
\hline
\multirow{2}{*}{\textbf{Average}} & P-SP &0.982&0.983&0.983&0.953&0.983&0.984&0.985 &0.951&0.950\\
                                          & USEE &0.981&0.977&0.982&0.952&0.982&0.983& 0.984&0.949&0.948\\     
\hline

\end{tabular}
\label{tab:backtranslation_similarity}
\end{table*}

\renewcommand{\arraystretch}{1.1}
\begin{table*}[t]
\centering
\caption{The similarity between the original AI-generated texts and combined back-translations}
\begin{tabular}{lccccccccc}
\hline
\rowcolor{Gray}\textbf{Sim.} & \textbf{News} & \textbf{Abst.}& \textbf{ELI-G}& \textbf{ELI-L}& \textbf{ELI-M}& \textbf{ELI-P}& \textbf{ELI-Y} & \textbf{R-G}& \textbf{R-L} \\
\hline
P-SP &0.951&0.960&0.953&0.951&0.949&0.952& 0.955&0.881&0.872\\
USEE &0.947&0.945&0.946&0.945&0.940&0.944& 0.947&0.861&0.861\\
\hline
\end{tabular}
\label{tab:combined_similarity}
\end{table*}

\subsection{Reducing impact of back-translation evasions}
Our proposed countermeasure is grounded in the ESAS metric \cite{ayoobi2024seeing}.
By facilitating the prioritization of entities within the vocabulary, the ESAS metric provides a framework for identifying the most critical entities that differentiate human-written texts from those generated by LLMs. 
The ESAS metric is computed as follows:

\begin{equation}
\begin{split}
E^{(AI\, vs.\, Human)}_{w_i} &= P(w_i)\Bigl(H(\mathcal{A})-H(\mathcal{A}|\mathcal{W}=w_i)\Bigl)\\
 &= \frac{N_i}{N} \Bigl(1 + \frac{N_{L,i}}{N_i} \log(\frac{N_{L,i}}{N_i})+\frac{N_{H,i}}{N_i} \log(\frac{N_{H,i}}{N_i}) \Bigl)
\end{split}
\end{equation}
where $P(w_i)$, $H(\mathcal{A})$, $H(\mathcal{A}|\mathcal{W}=w_i)$, $N$, $N_i$, $N_{L,i}$, and $N_{H,i}$ represent the likelihood of
occurrence of $w_i$ in a text, entropy of authorship, entropy of authorship conditioned on the presence of entity $w_i$ in the text, the size of the vocabulary, the frequency of entity $w_i$, its frequency in LLM-generated text, and its frequency in human written text, respectively.

To adapt ESAS to account for back-translation, we begin by separating texts into three groups: human-written (H), AI-generated (A), and back-translated (B). 
ESAS scores are then calculated by comparing two sets at a time, resulting in six possible scenarios: 1. \{H\} vs \{A\}, 2. \{H\} vs \{B\}, 3. \{H\} vs \{A,B\}, 4. \{A\} vs \{B\}, 5. \{A\} vs \{B,H\}, and 6. \{B\} vs \{A,H\}. 
The final score assigned to each entity is a weighted sum of the scores obtained from all six scenarios.
\begin{equation}
\begin{split}
 M\!E\!S\!A\!S_{w_i} &= \alpha_1 E^{(H\, vs.\, A)}_{w_i} + \alpha_2 E^{(H\, vs.\, B)}_{w_i} + \alpha_3 E^{(H\, vs.\, A,B)}_{w_i} \alpha_4 E^{(A\, vs.\, B)}_{w_i}+\alpha_5 E^{(A\, vs.\, B,H)}_{w_i}+\alpha_6 E^{(B\, vs.\,A,H)}_{w_i}
\end{split}
\end{equation}

Here, $E^{(X\, vs.\, Y)}_{w_i}$ represents the 
ESAS score for entity $w_i$, calculated when comparing the separation of texts between group $X$ and group $Y$.
We refer to our proposed method as modified ESAS (MESAS).
The first three scenarios aid in distinguishing human-written texts from AI-generated texts while increasing robustness against back-translation. 
Conversely, the last three scenarios undermine the detector’s performance in terms of both detection accuracy and resistance to back-translation. 
During validation, we set $\alpha_1{=}\alpha_2{=}\alpha_3{=}0.5$ and $\alpha_4{=}\alpha_5{=}\alpha_6{=}-0.5$. 
After ranking the entities based on the MESAS metric, we select the top $q$ entities to be used in the TF-IDF method.
A logistic regression (LR) model is then trained on the features produced by the TF-IDF method, outputting the probability of the text being AI-generated. 
A probability near zero indicates human authorship, while a value close to one suggests AI authorship.
We introduce two configurations for MESAS. The first, MESAS (Uni), leverages uni-grams as the entities in the ESAS method.
The second configuration, MESAS (Uni+Bi), employs both uni-grams and bi-grams as separate entities in ESAS.
The probabilities from two distinct LR classifiers, one for uni-grams and one for bi-grams, are combined by averaging their probabilities, forming an ensemble model.

\section{Result and Discussion}
Although enhancing an AI detector's capability to identify AI-generated text is the primary objective, it is essential for the detector to minimize false positives by not labeling human-authored content as AI.
To facilitate fair comparison across detectors employing varied probability thresholds, we maintain a fixed false positive rate (FPR) of $1\%$ and report the true positive rate (TPR).

\subsection{Impact of back-translation on current AI text detection systems}
To assess the robustness of existing AI text detection methods against our back-translation manipulation, we conduct experiments on nine detectors: six open-source models (RADAR \cite{hu2023radar}, LLMDet \cite{wu2023llmdet}, Likelihood \cite{gehrmann2019gltr}, Rank \cite{gehrmann2019gltr,ippolito2019automatic}, Log-Rank, and ESAS \cite{ayoobi2024seeing}) and three commercial detectors (Pangram \cite{emi2024technicalreportpangramaigenerated}, GPTZero \cite{gptzero}, and ZeroGPT \cite{zerogpt}).
RADAR employs an adversarial learning framework and utilizes two language models: one functioning as a paraphraser and the other as a detector.
During the training phase, the detector is optimized to differentiate between human-authored and AI-generated text, while the paraphraser model evolves to modify AI-generated text to elude detection.
The LLMDet operates in two distinct phases: dictionary compilation and text source detection. For the latter, the algorithm computes proxy perplexity scores for specific LLMs by leveraging next-token probabilities of salient n-grams as features. 
The text's origin is subsequently determined through an analysis of these LLM-dependent proxy perplexities.
The Likelihood, Rank, and Log-Rank methods are statistical approaches grounded in token probability analysis. 
In the Likelihood method, the model’s average token log probability is used to determine if a text is AI-generated. 
Rank and Log-Rank methods rely on the average rank or log-rank of tokens.
Texts exhibiting lower average rank or log-rank values are indicative of AI generation.
Table \ref{tab:other_methods} presents the TPRs before and after applying back-translation manipulation. 
Open-source methods are displayed above the thick line, while closed-source methods are shown below.
Owing to budgetary limitations, GPTZero and ZeroGPT analyses were confined to $200$ randomly sampled texts per dataset, whereas full datasets were utilized for all other methods. 
In instances where sample size limitations precluded fixing the FPR at $1\%$, the actual FPR is indicated in parenthetical superscript format alongside the TPR.
The dataset containing answers to Reddit questions generated by GPT-3.5 Turbo, Llama3, Mistral, Phi3, and Yi is represented by ELI-G, ELI-L, ELI-M, ELI-P, and ELI-Y, respectively.
Additionally, the review datasets produced by GPT4o and Llama3.1 are denoted as R-G and R-L, respectively, in the table.

The results reveal that the review datasets raise significant concerns about the robustness of six out of nine detectors when it comes to identifying short AI-generated text, even before the application of back-translation manipulation. 
LLMDet exhibits a bias towards classifying texts as AI-generated, resulting in substantially diminished TPRs across all datasets when maintaining a low FPR.
The implementation of back-translation leads to an average $54.3\%$ reduction in RADAR's TPR. 
While Likelihood, Rank, and Log-Rank methods demonstrate poor performance in detecting AI-generated texts within News, Abstract, and review datasets, the application of back-translation significantly reduces their TPRs in ELI datasets.
A reduction of $50\%$, $65.4\%$, and $52\%$ in the average TPR is observed for the Likelihood, Rank, and Log-Rank methods, respectively.
The outcomes for GPTZero and ZeroGPT indicate a lack of robustness against back-translation techniques.
For example, GPTZero's efficacy on the R-L dataset, in terms of TPR, decreases considerably from $0.65$ to $0.09$.
Similarly, ZeroGPT experiences a dramatic TPR reduction on the ELI-M dataset, decreasing from $0.98$ to $0.03$.
In comparison to other methods, ESAS and Pangram exhibit a degree of robustness, particularly for datasets with longer texts (News, Abstract, and ELIs).
However, back-translation manipulation can conceal certain AI-written reviews from detection. 
Specifically, the TPR for review datasets decreases by an average of $26.5\%$ for ESAS and $12\%$ for Pangram.

\renewcommand{\arraystretch}{1.2}
\begin{table*}[t]
\centering
\caption{Performance of current detection methods before and after applying back-translation in terms of TPR with the FPR fixed at $1\%$. In cases where the FPR could not be held at $1\%$, the FPR is presented in parentheses as a superscript next to the TPR value. Methods marked with an asterisk (*) are tested using a sampled version of the datasets.}
\resizebox{\columnwidth}{!}{%
\begin{tabular}{llccccccccc}
\hline
\rowcolor{Gray}\textbf{}&  & \textbf{News} & \textbf{Abst.}& \textbf{ELI-G}& \textbf{ELI-L}& \textbf{ELI-M}& \textbf{ELI-P}& \textbf{ELI-Y}& \textbf{R-G}& \textbf{R-L}  \\
\hline
\multirow{2}{*}{\textbf{RADAR} } & Before &0.299&0.541&0.716&0.483&0.602&0.512&0.371 &0.000&0.000\\
                                & After  &0.096&0.416&0.344&0.143&0.255&0.152&0.202&0.001&0.000\\
\hline
\multirow{2}{*}{\textbf{LLMDet} } & Before  &0.004&0.008&0.018&0.043&0.031&0.030&0.024&0.005&0.008\\
                                & After  &0.009&0.008&0.016&0.046&0.027&0.028&0.018&0.007&0.012\\                      
\hline
\multirow{2}{*}{\textbf{Likelihood}} & Before  &0.018&0.004&0.449&0.874&0.651&0.729&0.667&0.002&0.004\\
                                     & After  &0.084&0.004&0.258&0.442&0.331&0.375&0.280&0.003&0.011\\                              
\hline
\multirow{2}{*}{\textbf{Rank}} & Before  &0.121&0.114&0.490&0.654&0.577&0.553&0.526&0.000&0.000\\
                                & After  &0.045&0.004&0.166&0.268&0.204&0.185&0.147&0.000&0.001\\                                
\hline
\multirow{2}{*}{\textbf{Log-Rank}} & Before  &0.042&0.010&0.534&0.937&0.696&0.791&0.765&0.001&0.004\\
                                & After  &0.085&0.003&0.268&0.485&0.347&0.385&0.301&0.002&0.006\\                     
\hline

\multirow{2}{*}{\textbf{ESAS}} & Before  &0.839&0.969&0.960&0.956&0.949&0.952&0.907&0.869&0.763\\
                                                        & After  &0.706&0.934&0.932&0.897&0.909&0.908&0.869&0.601&0.599\\  
 
\specialrule{2pt}{0pt}{0pt}
\multirow{2}{*}{\textbf{Pangram} } & Before  &0.969&0.998&0.994&0.959&0.986&0.998&0.993&0.858&0.864\\
                                & After  &0.936&0.999&0.987&0.993&0.974&0.989&0.987&0.79&0.725\\                             
\hline
\multirow{2}{*}{\textbf{GPTZero* }} & Before  &0.97&1\textsuperscript{(0.02)}&0.88&1\textsuperscript{(0.05)}&1\textsuperscript{(0.1)}&1\textsuperscript{(0.05)}&0.99&1\textsuperscript{(0.05)}&0.65\\
                                & After  &0.42&0.82&0.71&0.97&0.96&1\textsuperscript{(0.05)}&0.97\textsuperscript{(0.02)}&0.43&0.09\\                              
\hline
\multirow{2}{*}{\textbf{ZeroGPT*}} & Before  &0.95\textsuperscript{(0.63)}&0.5\textsuperscript{(0.08)}&0.97\textsuperscript{(0.08)}&0.99\textsuperscript{(0.24)}&0.96\textsuperscript{(0.23)}&0.98\textsuperscript{(0.08)}&0.98\textsuperscript{(0.24)}&0.08&0.05\textsuperscript{(0.02)}\\
                                & After  &0.17\textsuperscript{(0.63)}&0\textsuperscript{(0.08)}&0.04\textsuperscript{(0.08)}&0.1\textsuperscript{(0.24)}&0.04\textsuperscript{(0.23)}&0.03\textsuperscript{(0.08)}&0.02\textsuperscript{(0.24)}&0&0\textsuperscript{(0.02)}\\                                
\hline

\end{tabular}
}
\label{tab:other_methods}
\end{table*}
\renewcommand{\arraystretch}{1.2}
\begin{table*}[t]
\centering
\caption{Performance of the proposed counteract method before and after applying back-translation in terms of TPR with the FPR fixed at $1\%$.}
\resizebox{0.85\columnwidth}{!}{%
\begin{tabular}{llccccccccc}
\hline
\rowcolor{Gray}\textbf{}&  & \textbf{News} & \textbf{Abst.}& \textbf{ELI-G}& \textbf{ELI-L}& \textbf{ELI-M}& \textbf{ELI-P}& \textbf{ELI-Y}& \textbf{R-G}& \textbf{R-L}  \\
\hline
\multirow{2}{*}{\textbf{MESAS (Uni)}} & Before &0.827&0.927&0.957&0.946&0.934&0.963&0.895&0.808&0.784\\
                               & After  &0.824&0.922&0.964&0.965&0.953&0.961&0.940&0.644&0.744\\
\hline
\multirow{2}{*}{\textbf{MESAS (Uni+Bi)}} & Before &0.980&0.989&0.988&0.972&0.975&0.991&0.921&0.959&0.912\\
& After  &0.960&0.982&0.988&0.987&0.979&0.989&0.958&0.872&0.811\\
\hline
\end{tabular}
}
\label{tab:counter}
\end{table*}
\subsection{Evaluation of the proposed countermeasure}
In our experimental design, we implement MESAS with $q {=} 4000$ for entity  selection (uni-grams or bi-grams).
An LR model is trained on TF-IDF features, restricted to a vocabulary containing the $4000$ entities with maximal MESAS scores. 
The FPR is fixed at $1\%$, and the TPR is reported.
Table \ref{tab:counter} shows the effectiveness of the proposed MESAS method in counteracting back-translation manipulation. 
MESAS (Uni) demonstrates robust resilience, with only a $1.54\%$ average TPR reduction after manipulation.
MESAS (Uni+Bi) shows comparable stability, experiencing a mere $1.85\%$ decrease in average TPR.
It is noteworthy that although the ensemble method (MESAS (Uni+Bi)) shows a marginally higher TPR reduction, its average TPR of $0.947$ after back-translation surpasses MESAS (Uni) at $0.88$, emphasizing the ensemble's enhanced detection capabilities.

Furthermore, beyond enhancing robustness to back-translation, in some cases, the TPR even improved.
This observation hints at the possibility of using back-translation to enhance detection accuracy, similar to its effect in data augmentation. 
We leave further exploration of this potential for future research.

\subsection{Ablation study}
We conduct an ablation study to evaluate the influence of intermediate languages and combiner on the performance of back-translation manipulation as a detection evasion technique.

\subsubsection{Evaluation of each intermediate language to evading detection}
We conduct an iterative exclusion process, removing one language at a time from the set of intermediate languages. 
The WER combiner subsequently integrates back-translated texts from the nine remaining languages. 
Table \ref{tab:excluding} presents the TPRs for each language exclusion scenario after subjecting the manipulated text to the ESAS detection method.
The symbol $\varnothing$ represents the baseline condition where all ten languages are included. 
A TPR exceeding $\varnothing$ indicates a positive contribution of the excluded language to the efficacy of back-translation manipulation in evading detection.
Conversely, a TPR below $\varnothing$ suggests that the excluded language could potentially be eliminated or substituted with a more effective alternative.

Analysis of the results reveals negligible deviation from the baseline,  indicating that the proposed method is robust to the choice of languages.
However, Japanese emerges as the most influential intermediate language. In 8 out of 9 datasets, the exclusion of Japanese yields increased TPR.
Japanese demonstrates the most significant impact across multiple datasets, including News, ELI-L, ELI-M, ELI-P (alongside Dutch and Korean), ELI-Y, and R-G, when compared to other languages. 
For the Abstract dataset, the exclusion of Korean most substantially impairs the manipulation efficacy, while for the R-L dataset, Chinese exclusion yields the highest impact.
German slightly outperforms other languages in its impact on the ELI-G dataset.

\renewcommand{\arraystretch}{1.1}
\begin{table*}[t]
\centering
\caption{The impact of removing individual intermediate languages on detection evasion in terms of TPR with fixed FPR at $1\%$. The symbol $\varnothing$ represents the baseline condition where all ten languages are included.}
\begin{tabular}{lccccccccc}
\hline
\rowcolor{Gray}\textbf{Excluded} & \textbf{News} & \textbf{Abst.}& \textbf{ELI-G}& \textbf{ELI-L}& \textbf{ELI-M}& \textbf{ELI-P}& \textbf{ELI-Y} & \textbf{R-G}& \textbf{R-L} \\
\hline
$\varnothing$  &0.706&0.934&0.932&0.897&0.909&0.908&0.869&0.601&0.599\\\hline
            PT &0.711&0.936&0.931&0.904&0.908&0.910&0.869&0.605&0.591\\\hline
            ES &0.704&0.927&0.932&0.900&0.907&0.905&0.873&0.600&0.605\\\hline
            FR &0.710&0.934&0.932&0.897&0.908&0.904&0.868&0.596&0.599\\\hline
            IT &0.706&0.936&0.931&0.898&0.909&0.910&0.869&0.611&0.599\\\hline
            ZH &0.710&0.932&0.933&0.904&0.911&0.907&0.871&0.624&0.615\\\hline
            NL &0.704&0.937&0.931&0.901&0.906&0.911&0.873&0.609&0.600\\\hline
            DA &0.703&0.933&0.933&0.898&0.909&0.907&0.871&0.609&0.603\\\hline
            JA &0.724&0.937&0.931&0.905&0.915&0.911&0.883&0.676&0.601\\\hline
            DE &0.707&0.937&0.935&0.904&0.909&0.908&0.873&0.611&0.604\\\hline
            KO &0.718&0.949&0.931&0.901&0.909&0.911&0.870&0.599&0.596\\\hline
\end{tabular}
\label{tab:excluding}
\end{table*}

\subsubsection{Effect of the number of selected intermediate languages}
We assess the proposed manipulation by varying the number of intermediate languages involved. 
Beginning with all 10 languages, we sequentially eliminate one language at a time, adhering to the following order: Portuguese, Spanish, French, Italian, Chinese, Dutch, Danish, Japanese, German, and Korean.
At each step, the WER combiner merges the back-translated texts from the remaining languages.
The resultant manipulated texts are then evaluated using ESAS to measure the corresponding TPR.

Figure 2 depicts the change in TPR relative to the baseline with all 10 languages.
$\Delta$TPR increments indicate a reduction in manipulation efficacy, bringing performance closer to the pre-back-translation manipulation state. 
For most datasets, $\Delta$TPR remains negligible until the exclusion of four languages.
This pattern implies that six languages may constitute a threshold for maintaining manipulation effectiveness.
Upon exclusion of the fifth language, datasets containing shorter AI-generated texts (specifically R-G and R-L) display a more pronounced $\Delta$TPR incline compared to other datasets.
The Abstract, ELI-G, and ELI-M datasets maintain near-constant TPR values throughout the language reduction process until a single language remains.
This phenomenon may be attributed to the inherent robustness of the ESAS method when applied to these specific datasets, implying that the introduction of additional intermediate languages fails to substantially influence the evasion of detection in these instances.
Therefore, a more sophisticated combining approach may be necessary to further improve detection evasion.

\begin{figure}[t]
    \centering
    \includegraphics[width=0.8\linewidth]{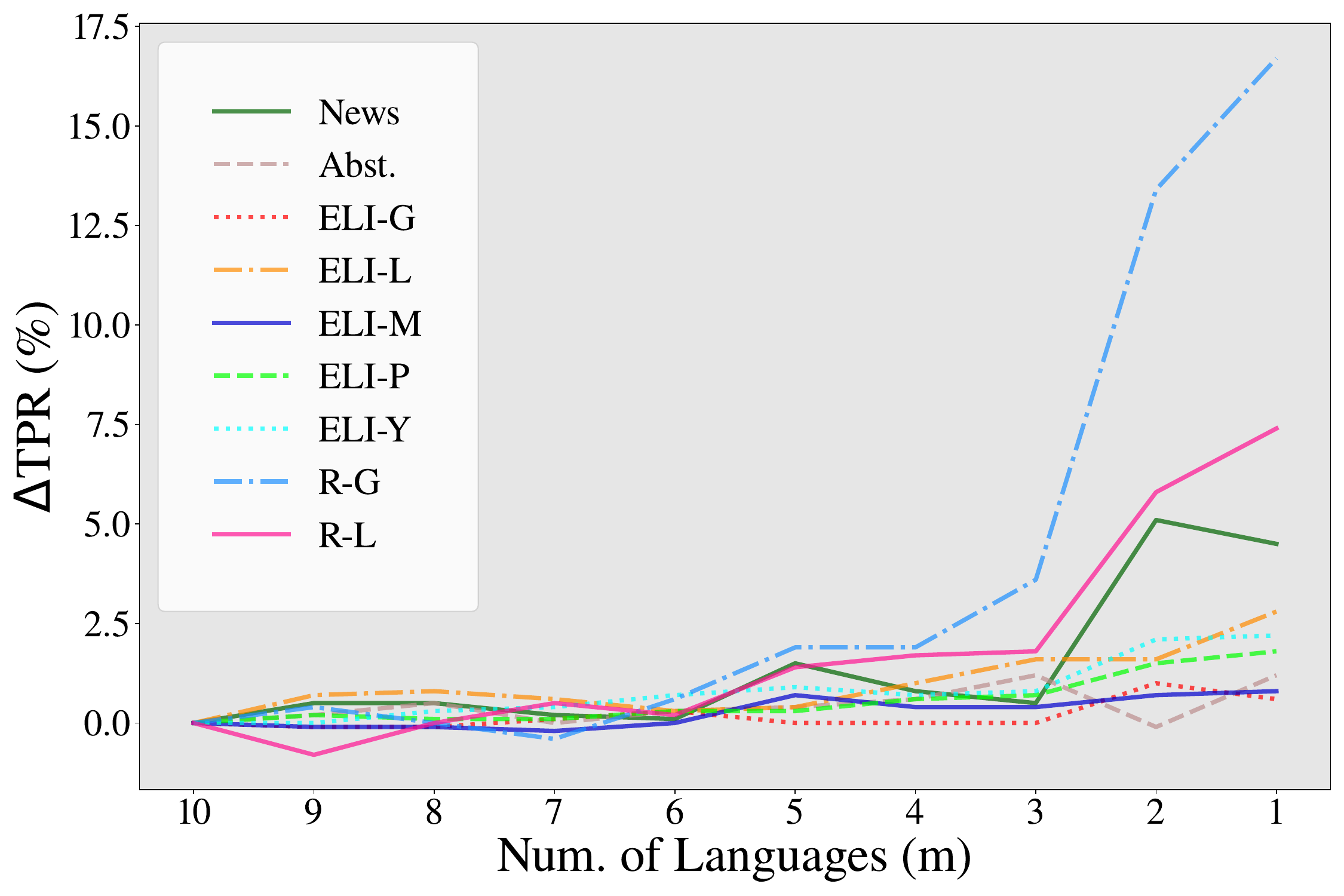}
    \caption{TPR variation from baseline (all 10 languages employed) with varying numbers of intermediate languages.}
    \label{fig:remove_increase}
\end{figure}

\subsubsection{Impact of combiner method on decreasing TPR}
\begin{figure}[t]
    \centering
    \includegraphics[width=\linewidth]{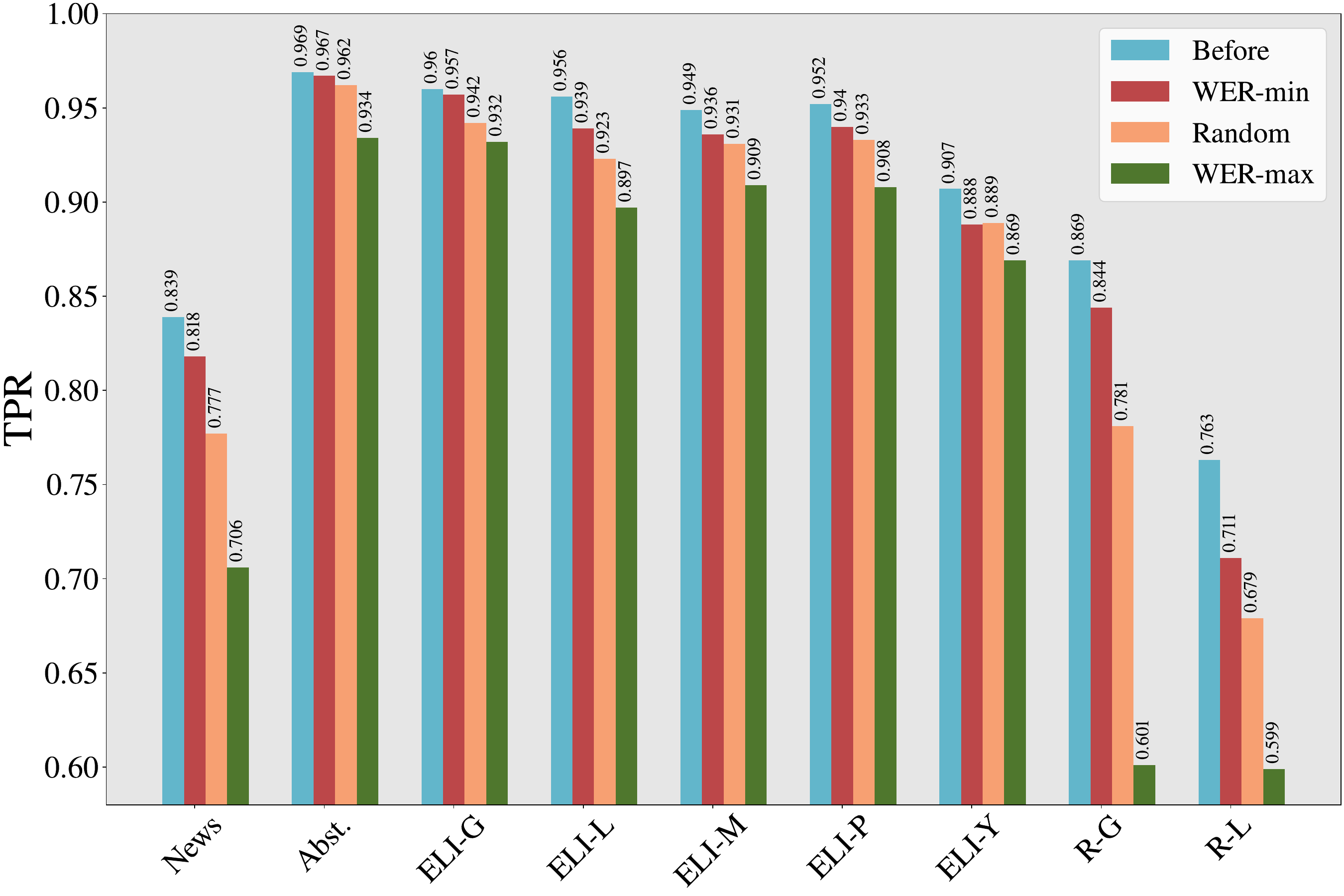}
    \caption{Comparison of different combiner methods in terms of TPR with fixed FPR at $1\%$.}
    \label{fig:random_min}
\end{figure}

We perform an experiment to evaluate the impact of the proposed WER-based combiner on evading ESAS detection.
The combiner method used throughout the paper, ``WER-max'', selects back-translated texts based on the maximum WER metric. 
For comparison, we develop ``WER-min'', which selects texts based on the lowest WER metric.
Additionally, a random combination approach, designated as ``Random'' is implemented, wherein original AI-generated sentences are replaced by randomly selected alternatives from the back-translated texts. 
Figure \ref{fig:random_min} presents a comparative bar plot illustrating the TPRs for the pre-manipulation baseline (denoted as ``Before''), ``Random'', ``WER-min'', and ``WER-max'' methodologies.

The results demonstrate that the proposed ``WER-max'' combiner consistently achieves lower TPRs compared to both the ``Random'' and ``WER-min'' methods across all datasets. 
``WER-min'' yields higher TPRs in 8 out of 9 cases, more closely resembling pre-manipulation TPRs compared to the ``Random'' method.
This outcome is consistent with the ``WER-min'' algorithm's selection criteria, which favor sentences with the smallest changes in terms of substitution, insertion, and deletion compared to the original AI text. 
Moreover, the efficacy of the ``Random'' method, which solely employs back-translated texts without additional processing, validates the inherent effectiveness of back-translation as a detection evasion technique.

\section{Conclusion and future work}
In this work, we highlight the concerning vulnerability in existing AI text detectors by introducing back-translation as an effective manipulation strategy to circumvent AI text detection.
Our findings demonstrate that this method preserves the semantic content of the original AI-generated text while significantly reducing the TPR of existing detectors. 
As a proactive defense against such exploits, we devised a detection mechanism that exhibits strong performance, experiencing only a $1.85\%$ drop in TPR following back-translation.
Furthermore, we contribute to the field by introducing a comprehensive dataset called ESPERANTO comprising texts in different writing styles and from 8 distinct LLMs, which has been made publicly accessible to support future research endeavors.

Our research was limited to an analysis of 10 preselected languages. Further studies are required to examine and rank additional languages, enabling the identification of superior candidates for the proposed back-translation manipulation technique.
In addition, future research should focus on developing more sophisticated combiner methods that incorporate additional linguistic features such as part-of-speech tags and grammatical structures. 
We hypothesize that such advanced techniques could further degrade the TPR of AI text detectors, presenting an avenue for subsequent investigation.

\section*{Ethical Considerations}
The intention of this study is to assess the robustness of current AI text detection algorithms.
The widespread use of LLMs and their potential for misuse, makes the robustness of AI text detectors essential for their role in investigative applications to combat AI-generated deceptions.
Any lack of robustness in these systems could lead to significant challenges in the future.
Therefore, this research is intended to assist detector developers in testing and validating their methodologies against potential manipulations. 
We emphasize that the findings presented herein should be utilized solely for assessment purposes and not for circumventing existing detection systems.

\section*{Acknowledgments}
The authors express their gratitude to Pangram Lab, with particular acknowledgment to Max Spero and Bradley Emi, for their provision of credits that facilitated the evaluation of our proposed method and datasets on the Pangram detector.

\bibliographystyle{unsrt}  
\bibliography{references}  
\end{document}